%% file: main.tex
\documentclass[letterpaper, 10 pt, conference]{ieeeconf}  

\IEEEoverridecommandlockouts                              

\usepackage{amsmath,amsfonts}
\usepackage{array}
\usepackage{stfloats}
\usepackage{url}
\usepackage{verbatim}
\usepackage{cite}
\def\BibTeX{{\rm B\kern-.05em{\sc i\kern-.025em b}\kern-.08em
    T\kern-.1667em\lower.7ex\hbox{E}\kern-.125emX}}
\usepackage{balance}

\usepackage{graphics} 
\usepackage{epsfig} 
\usepackage{mathptmx} 
\usepackage{multirow}
\usepackage{makecell}
\usepackage{xcolor}
\usepackage{algorithm}
\usepackage{algorithmicx}
\usepackage{comment}
\usepackage{algpseudocode}

\usepackage{footnote}
\usepackage[bottom]{footmisc}
\usepackage{threeparttable}
\usepackage{multirow}
\usepackage{tabularx}
\usepackage{ragged2e}
\usepackage{booktabs}
\usepackage{makecell}
\usepackage{hyperref}
\usepackage{xspace}
\usepackage{newtxtext, newtxmath}
\usepackage{soul}

\newcolumntype{C}[1]{>{\centering\arraybackslash}p{#1}}
\newcolumntype{P}[1]{>{\centering\arraybackslash}p{#1}}
\newcolumntype{M}[1]{>{\centering\arraybackslash}m{#1}}
\newcolumntype{L}[1]{>{\raggedright\arraybackslash}p{#1}}
\newcolumntype{R}[1]{>{\raggedleft\arraybackslash}p{#1}}
\newcolumntype{J}[1]{>{\justifying\arraybackslash}p{#1}}

\usepackage{footnote}
\usepackage[bottom]{footmisc}
\usepackage{threeparttable}
\usepackage{tabularx}
\usepackage{ragged2e}
\usepackage{hyperref}

\usepackage{xspace}
\newcommand{\mypara}[1]{\noindent\textbf{#1}}

\definecolor{amethyst}{rgb}{0.6, 0.4, 0.8}
\definecolor{aqua}{rgb}{0.0, 1.0, 1.0}
\definecolor{arylideyellow}{rgb}{0.91, 0.84, 0.42}
\definecolor{phlox}{rgb}{0.87, 0.0, 1.0}
\definecolor{caribbeangreen}{rgb}{0.0, 0.8, 0.6}

\title{\LARGE \bf Contextual Client Selection \\for Federated Learning in Vehicular Networks}
\title{\LARGE \bf V2X-Boosted Federated Learning for Cooperative Intelligent Transportation Systems with Contextual Client Selection}

\author{Rui Song$^{1,2}$, Lingjuan Lyu$^{3}$, Wei Jiang$^{4}$, Andreas Festag$^{1,5}$ and Alois Knoll$^{2}$
\thanks{This work was supported by the German Federal Ministry for Digital and Transport (BMVI) in the projects ``KIVI -- KI im Verkehr Ingolstadt'' and ''5GoIng – 5G Innovation Concept Ingolstadt''.}%
\thanks{$^{1}$Rui Song and Andreas Festag are with Fraunhofer Institute for Transportation and Infrastructure Systems IVI, Ingolstadt, Germany, e-mail:
        {\tt\small \{rui.song, andreas.festag\}@ivi.fraunhofer.de}.}%
\thanks{$^{2}$Rui Song and Alois Knoll are with Technical University of Munich, Robotics, Artificial Intelligence and Real-Time Systems, Garching, Germany, e-mail:~{\tt\small rui.song@tum.de, knoll@in.tum.de}.}%
\thanks{$^{3}$Lingjuan~Lyu is with Sony AI, Tokyo, Japan, \mbox{e-mail:}~{\tt\small lingjuan.lv@sony.com}.}%
\thanks{$^{4}$Wei~Jiang is with German Research Center for Artificial Intelligence (DFKI), Kaiserslautern, Germany, e-mail:~{\tt\small wei.jiang@dfki.de}.}%
\thanks{$^{5}$Andreas Festag is with Technische Hochschule Ingolstadt, CARISSMA Institute for Electric, COnnected, and Secure Mobility (C-ECOS), Ingolstadt, Germany, e-mail:~{\tt\small andreas.festag@carissma.eu}.}%
}
\begin{document}

\maketitle

\input{sections/00_abstract}
\input{sections/01_intro}
\input{sections/02_related_work.tex}

\input{sections/04_method}
\input{sections/06_experiment}
\input{sections/07_conclusion}



\bibliography{ref}
\bibliographystyle{IEEEtran}

\newpage
\onecolumn
\input{sections/08_appendix.tex}
\end{document}

%% file: sections/00_abstract.tex
\begin{abstract}
Machine learning (ML) has revolutionized transportation systems, enabling autonomous driving and smart traffic services. Federated learning (FL) overcomes privacy constraints by training ML models in distributed systems, exchanging model parameters instead of raw data. However, the dynamic states of connected vehicles affect the network connection quality and influence the FL performance. To tackle this challenge, we propose a contextual client selection pipeline that uses Vehicle-to-Everything (V2X) messages to select clients based on the predicted communication latency. The pipeline includes: (i)~fusing V2X messages, (ii)~predicting future traffic topology, (iii)~pre-clustering clients based on local data distribution similarity, and (iv)~selecting clients with minimal latency for future model aggregation. Experiments show that our pipeline outperforms baselines on various datasets, particularly in non-iid settings. 
\end{abstract}

%% file: sections/01_intro.tex
\section{Introduction}
\label{sec:intro}

Machine learning (ML), a subfield of artificial intelligence, focuses on developing learning algorithms and inference models that enable digital systems to make decisions and predictions in terms of the knowledge learned from data. Over the past years, ML-based approaches exhibited great potential to revolutionize various scientific, engineering, economic, and cultural fields with outstanding technological advancements such as Google AlphaGo and Open AI’s ChatGPT. In the filed of road transportation, ML is possible to empower numerous new applications for realizing Intelligent Transportation System (ITS), e.g., environmental perception, road traffic flow optimization, and trajectory planning, which can significantly enhance the safety and efficiency of transportation systems~\cite{hu2023collaboration, xuv2xvit, li2023among, lei2022latency, xu2022cobevt}.



\begin{figure}[t]
\centering
\includegraphics[width=0.5\textwidth]{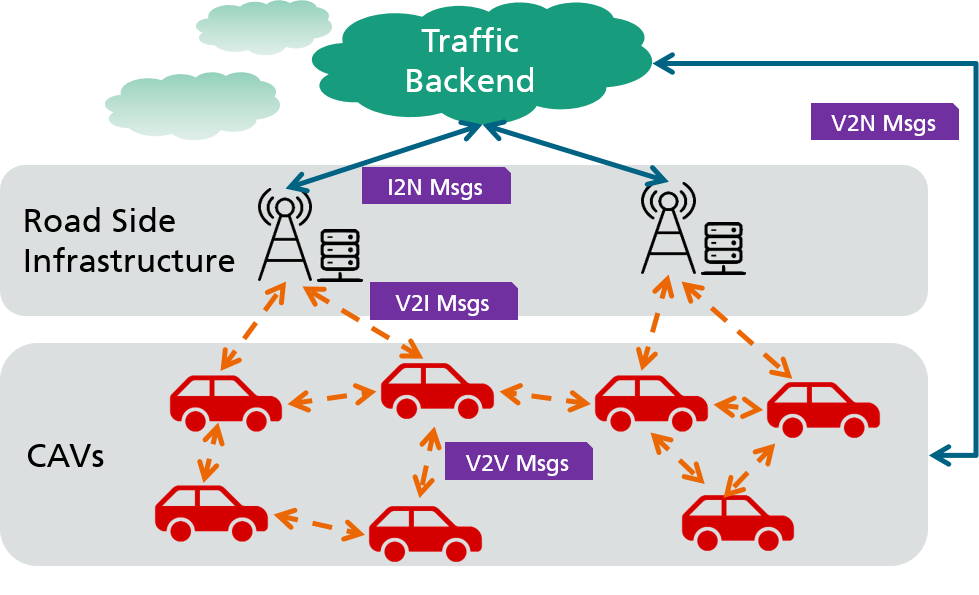}
\caption{An overview of vehicular networks in C-ITS, including vehicle-to-vehicle (V2V), vehicle-to-infrastructure (V2I), vehicle-to-network (V2N), and infrastructure-to-network (I2N) communication.}
\label{fig:v2x}
\end{figure}

Recently, a new ITS concept referred to as Cooperative
Intelligent Transportation System (C-ITS) attacked a lot
of interests from both academia and industry~\cite{sjoberg2017_CITS}. In C-ITS, the cooperation between two or more ITS sub-systems
(personal, vehicle, roadside and central) offers better quality and an enhanced service level, compared to that of the conventional ITS. As illustrated in Fig.~\ref{fig:v2x}, road participants –- specifically, connected automated vehicles (CAVs) –- can share information with one another through vehicle-to-everything (V2X) networks, which encompass vehicle-to-vehicle (V2V), vehicle-to-infrastructure (V2I), vehicle-to-network(V2N), and infrastructure-to-network (I2N).

The European standards for C-ITS define several types of V2X messages to facilitate decentralized information sharing. Specifically, for cooperative awareness and perception, dedicated message types -- the Cooperative Awareness Message (CAM) and the Collective Perception Message (CPM)\footnote{European Telecommunications Standards Institute (ETSI) \url{http://etsi.org/standards}, specifically EN 302 637-2 for the cooperative awareness service and TS 103 324 for the collective perception service.} -- are periodically exchanged among CAVs and with roadside infrastructure\cite{thandavarayan2020}. By sending and receiving V2X messages, enriched and improved environmental data of road traffic can be made available within vehicular networks.

In centralized model training using ML, CAV clients transmit data to a centralized system through vehicle-to-network communications. This process can generate an enormous volume of data, potentially exceeding the network's capacity. Moreover, data collected from CAV clients for ML model training cannot be directly shared due to privacy concerns. Differing from conventional ML, federated learning (FL) trains ML models using data from distributed systems, such as devices or clients, without centralizing the data~\cite{9354925}. In FL, connected clients share a model trained on their local data with a server, which aggregates the local models and updates the global model. The updated global model is then shared back with the clients. This process is repeated for a sufficient number of communication rounds until FL converges.

The deployment of 5G-V2X vehicular networks has further facilitated the use of FL in C-ITS by providing higher data rates and greater reliability for data exchange. This allows for the training of larger ML models for C-ITS applications and services, such as~\cite{ li2023voxformer,li2023multi,Where2comm22}. Although FL has great potential to preserve privacy and utilize a broader range of data resources~\cite{9360666}, the employment of FL in C-ITS has to address major challenges due to heterogeneity in data and networks, which can not only limit the performance but also lead to FL failures.


    \mypara{Data heterogeneity.} Data across clients is non-iid (non-identically independently distributed), resulting from various sensor types, combinations, poses, road scenes, traffic scenarios, climate and weather conditions, and more.
    
    \mypara{Network heterogeneity.} The diverse connection qualities of clients can slow down model sharing and cause communication delays for global model aggregation, which impedes the FL process.

To address these challenges and enhance the application of FL in C-ITS, we propose a novel FL framework. The main idea is to select clients for upcoming communication rounds based on (\emph{i}) the prediction of connection qualities in the context of road traffic status and (\emph{ii}) the similarity of local data distribution in clients. 


%% file: sections/02_related_work.tex
\section{Background and Related Work}
\label{sec:related_work}

\begin{figure*}[t!]
\centering
\includegraphics[width=1\textwidth]{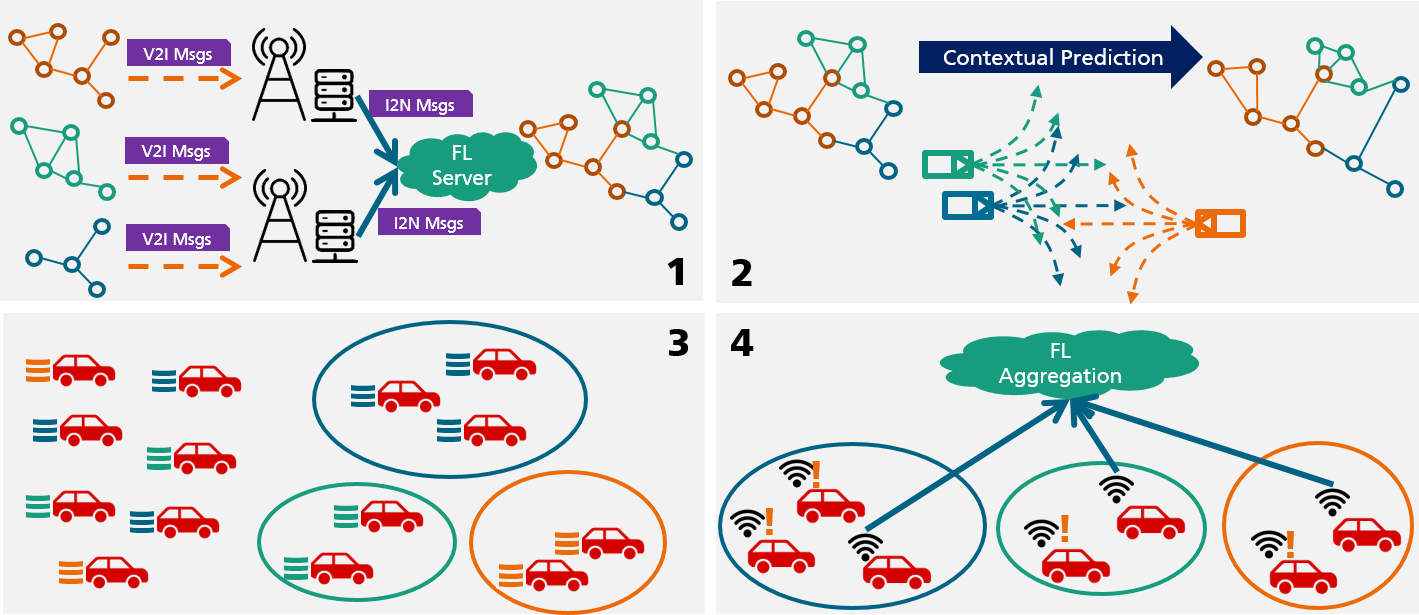}
\caption{Contextual client selection pipeline: (\textbf{1})~V2X message fusion; (\textbf{2})~Road traffic topology graph (RTTG) prediction; (\textbf{3})~Data-level client grouping; (\textbf{4})~Network-level client selection.}
\label{fig:system}
\end{figure*}

We discuss greedy and gossip client selection in FL, data- and network-based strategies, and FL in vehicular networks considering road traffic features.

\mypara{Greedy and gossip client selection.} 
FL, initially proposed by McMahan et al.~\cite{mcmahan2017communication}, suffers from the straggler effect due to varying connection qualities~\cite{jin2022accelerated}. Greedy client selection includes all clients in each communication round, while gossip (stochastic greedy) selection randomly selects connected clients. Both strategies struggle to avoid the straggler effect.

\mypara{Network-based client selection.}
Strategies focusing on network quality~\cite{8761315, abdulrahman2020fedmccs, chahoud2023feasibility, xu2020client} reduce the straggler effect but aren't specifically designed for vehicular networks with dynamic connection qualities and high-priority traffic services. Inspired by~\cite{fu2022digital}, we optimize client selection by predicting communication latency in vehicular networks.

\mypara{Data-based client selection.}
Client selection based on data distribution tackles heterogeneity. The approaches in~\cite{yin2018gradient,9345723,cho2022towards,shyn2022empirical, balakrishnan2021diverse, shen2022fast} consider data heterogeneity but overlook network parameters. Our work addresses both data distribution and network quality in vehicular networks. A comparison of the client selection paradigms is shown in Tab.~\ref{table:comparison}.

%% file: sections/04_method.tex
\section{Framework}
\label{sec:framework}

Our \emph{contextual client selection} framework is illustrated in Fig.~\ref{fig:system}, comprising V2X information sharing, traffic topology prediction, data-level client clustering, and network-level client clustering.

\mypara{V2X message fusion.}
We first fuse V2X messages. Continuously receiving CAM and CPM enables dynamic road maps with traffic object states. Road-side infrastructure collects and forwards V2X messages to a server via V2I and I2N networks. The server filters and fuses messages, obtaining traffic object states, such as position, speed, and acceleration. Fused results form an road traffic topology graph (RTTG), with each CAV characterized by a node with attributes. The RTTG digitizes C-ITS and recreates vehicular networks virtually.

\mypara{RTTG prediction.}
We predict future RTTGs. After V2X message fusion, we initialize a prediction instance for each CAV to estimate its trajectory. Predicted trajectories build future RTTGs. Predicted RTTGs integrate with digital C-ITS, providing possible connection quality for each CAV. We simulate networks in digital twin and calculate FL communication latency based on predictive transport scenarios.

\mypara{Data-level client grouping.}
We cluster clients into groups considering data heterogeneity. Our goal is to group clients with similar data distribution, ensuring each subset represents the whole group's data features. We observe model updates, considering gradient similarity as a data similarity criterion~\cite{yin2018gradient}. We group clients based on model parameter similarity. Clients must report gradient updates before a deadline for inclusion in data-level client grouping. After grouping, each subset represents its cluster. Selecting at least one client per cluster ensures satisfactory training performance.

\mypara{Network-level client election.}
We elect clients in each group based on contextual communication latency. Using predicted RTTG latency, we determine efficient client contributions for upcoming communication rounds. We employ the \emph{Fast-$\gamma$} rule, selecting the $\gamma$ clients with the lowest communication delay ($0<\gamma<1$) per cluster.

Through these stages, representative clients with minimal contextual communication latency are chosen for model aggregation. This process increases FL communication efficiency by optimizing communication rounds and round duration. De-selected clients save computational resources by not training models locally.

%% file: sections/06_experiment.tex
\section{Performance evaluation}
\label{sec:experiment}

\begin{figure*}[ht!]
   \centering
   \includegraphics[width=0.9\textwidth]{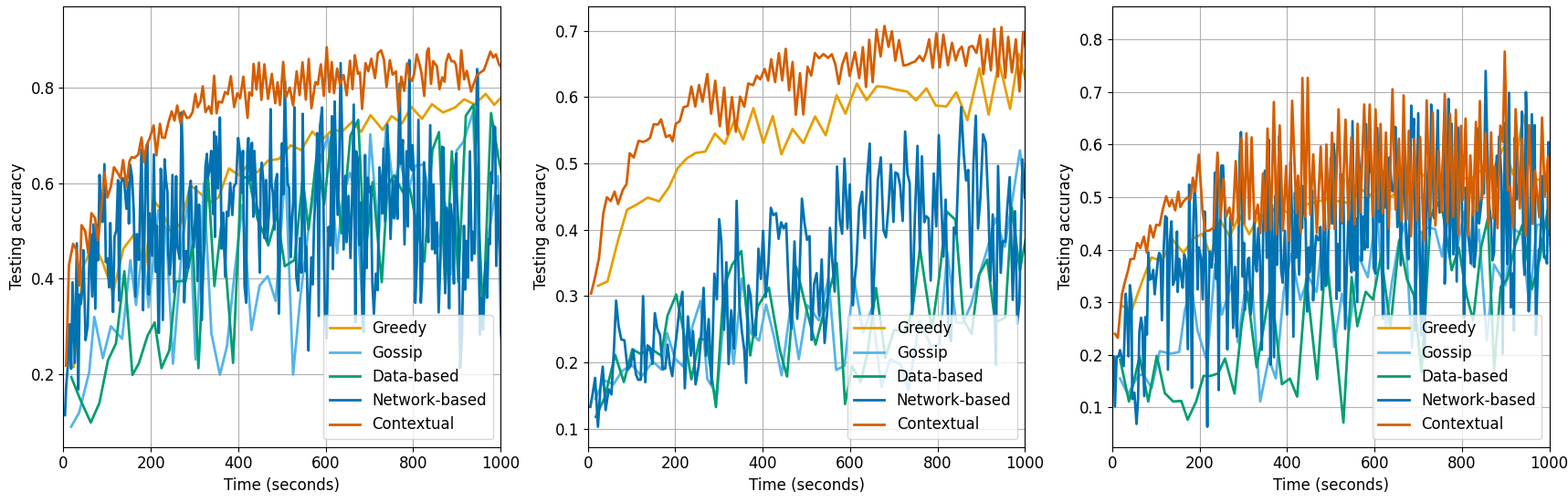}
   \caption{FL training performance (testing accuracy changes over time in seconds) using various client selection strategies on non-iid MNIST (left), CIFAR-10 (middle) and SVHN (right) data distributed in 100~clients.}
   \label{fig:performance}
\end{figure*}

\begin{table}[t]
\vspace{2mm}
\centering 
\begin{threeparttable}
\caption{\centering Required time to reach 0.5 of test accuracy using various paradigms of client selection strategies with decreasing connection rate (CR).}
\label{table:comparison_cr}
\begin{tabular}{M{0.5cm}L{2cm}L{1.8cm}L{2cm}}
   \toprule 
    \textbf{CR} & \textbf{Strategy} & \textbf{Time (s)} & \textbf{Reduction rate}\\
    \midrule 
    \midrule 
   \multirow{1}{*}{-} & Gossip & 3\,891.14 & 1$\times$\\
   \midrule 
    \multirow{3}{*}{1.0} & Data-based & 213.50 & 18.23$\times$\\
    & Network-based & 620.47 & 6.27$\times$\\
    & Contextual & l93.77 & \textbf{20.08$\times$}\\
    \midrule 
   \multirow{3}{*}{0.5} & Data-based & 2\,446.75 & 1.59$\times$\\
    & Network-based & 690.29 & 5.64$\times$\\
    & Contextual & l79.54 & \textbf{21.67$\times$}\\
    \midrule 
   \multirow{3}{*}{0.2} & Data-based & 2\,563.20 & 1.52$\times$\\
    & Network-based & 634.12 & 6.14$\times$\\
    & Contextual & l86.47 & \textbf{20.87$\times$}\\

    \bottomrule 
\end{tabular}
\end{threeparttable}
\end{table}

We implement and demonstrate FL with our pipeline as well as other four baselines, i.e. greedy, gossip, data-based and network-based client selection strategy, on a computer cluster with 4$\times$~NVIDIA-A100-PCIE-40GB GPUs and 4$\times$~32-Core-AMD-EPYC-7513 CPUs. The environment is a Linux system with Pytorch 1.8.1 and Cuda 11.1.

\subsection{Experiment setup}

We conduct the experimental evaluation by training models on three widely used open datasets MNIST~\cite{mnist-2010}, \mbox{CIFAR-10}~\cite{cifar10} and SVHN~\cite{netzer2011reading} distributed into 100 CAV clients in non-iid setting.\footnote{In our default non-iid setting, each client owns only 2 out of 10 classes.}

We compare our pipeline with four other client selection strategies as baselines, i.e., greedy, gossip, data-based and network-based, as described in Sec.~\ref{sec:related_work}. The learning rate is 0.001 and the batch size is 64. The number of the local epochs is set as 3 for training on MNIST, and 1 for training on CIFAR-10 and SVHN, respectively. Except the greedy strategy (all clients are selected in each communication round), the general selection rate for FL clients is defined as 10\%, i.e. around 10 clients are selected in each communication round.

\subsection{Performance results}

We show the general performance of FL with contextual client selection for training models on three datasets distributed in 100~vehicle clients with respect to default non-iid setting. We train deep learning models with different sizes on MNIST, CIFAR-10 and SVHN as FL tasks. As the experimental results in Fig.~\ref{fig:performance} show for all three tasks, FL with our contextual client selection can outperform the other four baselines. Generally, the FL with contextual client selection can achieve remarkable higher test accuracy than the other four strategies. Even though the network-based strategy allows the ML-model to be trained to a comparable test accuracy on SVHN, the contextual client selection results showcases much more stable convergence, as the data heterogeneity across CAVs are taken into account.

We conduct the experiments with various connection rates and evaluate the performance of FL. We take the required time to reach 0.5 of test accuracy for FL with gossip client selection as a baseline, and evaluate the time reduction rate of FL with other strategies. As the comparison results show in Tab.~\ref{table:comparison_cr}, FL with contextual client selection always needs less time than other two strategies at each connection rate. The time reduction rates are robustly over 20$\times$ even when only 20\% of clients are connected in networks. 

%% file: sections/07_conclusion.tex
\section{Conclusion}
\label{sec:conclusion}

In this work, we reviewed the existing client selection strategies for FL and introduced a novel four-stage V2X-Boosted FL pipeline for C-ITS.
The approach tackles both data and network heterogeneity in vehicular networks, boosting communication efficiency by reducing the number of communication rounds and shortening the time required for each round.
Compared to other strategies, FL with contextual client selection achieves higher accuracy and more stable convergence performance by leveraging V2X messages disseminated in vehicular networks. Future work will further consider the analytical model of communication networks and conduct more validation in traffic scenario data, such as~\cite{xu2022opv2v, li2022v2x, xu2023v2v4real}.

%% file: sections/08_appendix.tex
\section{Appendix}

\begin{table*}[h]
\vspace{2mm}
\centering 
\begin{threeparttable}
\caption{\centering Comparison of various paradigms of client selection strategies.}
\label{table:comparison}
\begin{tabular}{L{3cm}L{3cm}L{3cm}L{3cm}L{3cm}}
   \toprule 
    \textbf{Greedy}  & \textbf{Gossip} & \textbf{Data-based} & \textbf{Network-based} & \textbf{Contextual (ours)}\\
    \midrule 
   \vspace{1mm}
    All clients should be selected in each FL communication round. & A random subset of all connected clients is selected in each FL communication round. & Clients are selected according to the similarity of local data distribution.  &  Clients are selected according to the connection quality and availability in networks. & Clients are selected in consideration of both data and network heterogeneity.\\
    \bottomrule 
\end{tabular}
\end{threeparttable}
\end{table*}

\begin{figure*}[h]
   \centering
   \includegraphics[width=1\textwidth]{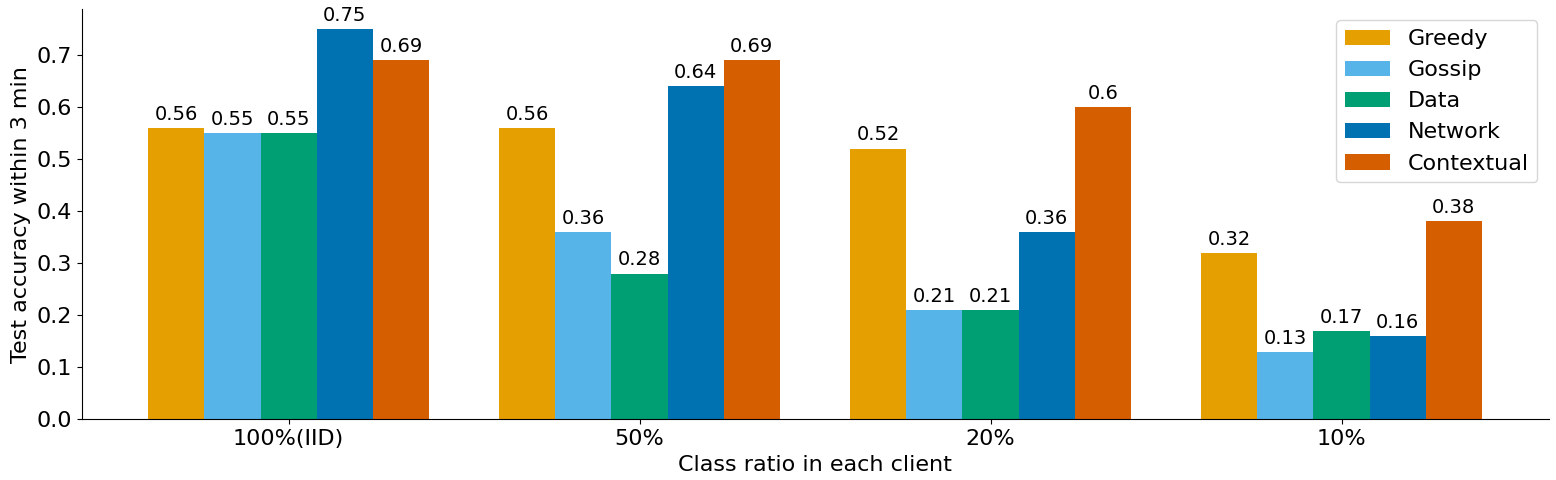}
   \caption{Test accuracy of FL with baselines and contextual client selection on CIFAR-10 data distributed in different class ratio in each client.}
   \label{fig:niid}
\end{figure*}

We also demonstrate 
various client selection strategies under different class ratio in each client to evaluate the performance in non-iid settings. We consider a scenario for training a ML-model with 100~CAVs for three minutes. Note that the class ratio 100\% indicates iid setting. As Fig.~\ref{fig:niid} shows, the pure network-based strategy can make the test accuracy of FL higher than others, because the data heterogeneity is not needed to be considered under ideal iid settings. However, in non-iid settings, the contextual client selection enhances FL and leads to a better test accuracy than other strategies. For instances, when the class ratio is 20\%, it achieves 2.85$\times$ test accuracy compared to a pure data-based and 1.67$\times$ to a pure network-based client selection strategy. Even in extremely non-iid setting with only 1 class in each client, the contextual client selection can reach 38\% test accuracy within three minutes, while the FL with network- or data-based strategies cannot converge.

%% file: main.bbl
\begin{thebibliography}{10}
\providecommand{\url}[1]{#1}
\csname url@samestyle\endcsname
\providecommand{\newblock}{\relax}
\providecommand{\bibinfo}[2]{#2}
\providecommand{\BIBentrySTDinterwordspacing}{\spaceskip=0pt\relax}
\providecommand{\BIBentryALTinterwordstretchfactor}{4}
\providecommand{\BIBentryALTinterwordspacing}{\spaceskip=\fontdimen2\font plus
\BIBentryALTinterwordstretchfactor\fontdimen3\font minus
  \fontdimen4\font\relax}
\providecommand{\BIBforeignlanguage}[2]{{%
\expandafter\ifx\csname l@#1\endcsname\relax
\typeout{** WARNING: IEEEtran.bst: No hyphenation pattern has been}%
\typeout{** loaded for the language `#1'. Using the pattern for}%
\typeout{** the default language instead.}%
\else
\language=\csname l@#1\endcsname
\fi
#2}}
\providecommand{\BIBdecl}{\relax}
\BIBdecl

\bibitem{hu2023collaboration}
Y.~Hu \emph{et~al.}, ``Collaboration helps camera overtake {LiDAR} in {3D}
  detection,'' in \emph{IEEE Conference on Computer Vision and Pattern
  Recognition (CVPR)}, 2023, doi:~10.48550/arXiv.2303.13560.

\bibitem{xuv2xvit}
R.~Xu, H.~Xiang, Z.~Tu, X.~Xia, M.-H. Yang, and J.~Ma, ``{V2X-ViT}:
  Vehicle-to-everything cooperative perception with vision transformer,'' in
  \emph{European Conference on Computer Vision (ECCV 2022)}, 2022,
  doi:~10.1007/978-3-031-19842-7{\_7}.

\bibitem{li2023among}
Y.~Li, Q.~Fang, J.~Bai, S.~Chen, F.~Juefei-Xu, and C.~Feng, ``Among us:
  Adversarially robust collaborative perception by consensus,'' \emph{arXiv
  preprint arXiv:2303.09495}, 2023.

\bibitem{lei2022latency}
Z.~Lei, S.~Ren, Y.~Hu, W.~Zhang, and S.~Chen, ``Latency-aware collaborative
  perception,'' in \emph{European Conference on Computer Vision (EECCV}, 2022,
  doi:~10.1007/978-3-031-19824-3{\_19}.

\bibitem{xu2022cobevt}
R.~Xu \emph{et~al.}, ``{CoBEVT}: Cooperative bird's eye view semantic
  segmentation with sparse transformers,'' in \emph{Conference on Robot
  Learning (CoRL)}, 2022, doi:~10.48550/arXiv.2207.02202.

\bibitem{sjoberg2017_CITS}
K.~Sj\"oberg, P.~Andres, T.~Buburuzan, and A.~Brakemeier, ``Cooperative
  {Intelligent} {Transport} {Systems} in {Europe}: Current deployment status
  and outlook,'' \emph{IEEE Vehicular Technology Magazine}, vol.~12, no.~2, pp.
  89--97, 2017, doi:~10.1109/MVT.2017.2670018.

\bibitem{thandavarayan2020}
G.~{Thandavarayan}, {M. Sepulcre}, and J.~{Gozalvez}, ``Generation of
  cooperative perception messages for connected and automated vehicles,''
  \emph{IEEE Transactions on Vehicular Technology}, vol.~69, no.~12, pp.
  16\,336--16\,341, 2020, doi:~10.1109/TVT.2020.3036165.

\bibitem{9354925}
R.~Yu and P.~Li, ``Toward resource-efficient federated learning in mobile edge
  computing,'' \emph{IEEE Network}, vol.~35, no.~1, pp. 148--155, 2021,
  doi:~10.1109/MNET.011.2000295.

\bibitem{li2023voxformer}
Y.~Li \emph{et~al.}, ``{VoxFormer}: Sparse voxel transformer for camera-based
  {3D} semantic scene completion,'' in \emph{IEEE/CVF Computer Vision and
  Pattern Recognition Conference (CVPR)}, 2023, doi:~10.48550/arXiv.2302.12251.

\bibitem{li2023multi}
Y.~Li, J.~Zhang, D.~Ma, Y.~Wang, and C.~Feng, ``Multi-robot scene completion:
  Towards task-agnostic collaborative perception,'' in \emph{Conference on
  Robot Learning}.\hskip 1em plus 0.5em minus 0.4em\relax PMLR, 2023, pp.
  2062--2072, url:~\url{https://proceedings.mlr.press/v205/li23e/li23e.pdf}.

\bibitem{Where2comm22}
Y.~Hu, S.~Fang, Z.~Lei, Z.~Yiqi, and C.~Siheng, ``Where2comm:
  Communication-efficient collaborative perception via spatial confidence
  maps,'' in \emph{Conference on Neural Information Processing Systems
  (NeurIPS)}, 2022, doi:~10.48550/arXiv.2209.12836.

\bibitem{9360666}
J.~Posner, L.~Tseng, M.~Aloqaily, and Y.~Jararweh, ``Federated learning in
  vehicular networks: Opportunities and solutions,'' \emph{IEEE Network},
  vol.~35, no.~2, pp. 152--159, 2021, doi:~10.1109/MNET.011.2000430.

\bibitem{mcmahan2017communication}
B.~McMahan, E.~Moore, D.~Ramage, S.~Hampson, and B.~A. y~Arcas,
  ``Communication-efficient learning of deep networks from decentralized
  data,'' in \emph{Artificial intelligence and statistics}.\hskip 1em plus
  0.5em minus 0.4em\relax PMLR, 2017, pp. 1273--1282,
  url~\url{https://proceedings.mlr.press/v54/mcmahan17a/mcmahan17a.pdf},.

\bibitem{jin2022accelerated}
J.~Jin, J.~Ren, Y.~Zhou, L.~Lyu, J.~Liu, and D.~Dou, ``Accelerated federated
  learning with decoupled adaptive optimization,'' in \emph{International
  Conference on Machine Learning}.\hskip 1em plus 0.5em minus 0.4em\relax PMLR,
  2022, pp. 10\,298--10\,322.

\bibitem{8761315}
T.~Nishio and R.~Yonetani, ``Client selection for federated learning with
  heterogeneous resources in mobile edge,'' in \emph{2019 IEEE International
  Conference on Communications (ICC)}, 2019, pp. 1--7,
  doi:~10.1109/ICC.2019.8761315.

\bibitem{abdulrahman2020fedmccs}
S.~AbdulRahman, H.~Tout, A.~Mourad, and C.~Talhi, ``{FedMCCS}: Multicriteria
  client selection model for optimal iot federated learning,'' \emph{IEEE
  Internet of Things Journal}, vol.~8, no.~6, pp. 4723--4735, 2020,
  doi:~10.1109/JIOT.2020.3028742.

\bibitem{chahoud2023feasibility}
M.~Chahoud, S.~Otoum, and A.~Mourad, ``On the feasibility of federated learning
  towards on-demand client deployment at the edge,'' \emph{Information
  Processing \& Management}, vol.~60, no.~1, p.~10, 2023,
  doi:~10.1016/j.ipm.2022.103150.

\bibitem{xu2020client}
J.~Xu and H.~Wang, ``Client selection and bandwidth allocation in wireless
  federated learning networks: A long-term perspective,'' \emph{IEEE
  Transactions on Wireless Communications}, vol.~20, no.~2, pp. 1188--1200,
  2020, doi:~10.1109/TWC.2020.3031503.

\bibitem{fu2022digital}
Y.~Fu \emph{et~al.}, ``Digital twin based network latency prediction in
  vehicular networks,'' \emph{MDPI Electronics}, vol.~11, no.~14, p.~21, 2022,
  doi:~10.3390/electronics11142217.

\bibitem{yin2018gradient}
D.~Yin \emph{et~al.}, ``Gradient diversity: a key ingredient for scalable
  distributed learning,'' in \emph{International Conference on Artificial
  Intelligence and Statistics}.\hskip 1em plus 0.5em minus 0.4em\relax PMLR,
  2018, pp. 1998--2007,
  url~\url{http://proceedings.mlr.press/v84/yin18a/yin18a.pdf},.

\bibitem{9345723}
W.~Zhang, X.~Wang, P.~Zhou, W.~Wu, and X.~Zhang, ``Client selection for
  federated learning with non-{IID} data in mobile edge computing,'' \emph{IEEE
  Access}, vol.~9, pp. 24\,462--24\,474, 2021,
  doi:~10.1109/ACCESS.2021.3056919.

\bibitem{cho2022towards}
Y.~J. Cho, J.~Wang, and G.~Joshi, ``Towards understanding biased client
  selection in federated learning,'' in \emph{International Conference on
  Artificial Intelligence and Statistics}.\hskip 1em plus 0.5em minus
  0.4em\relax PMLR, 2022, pp. 10\,351--10\,375,
  url:~\url{https://proceedings.mlr.press/v151/jee-cho22a.html},.

\bibitem{shyn2022empirical}
S.~K. Shyn, D.~Kim, and K.~Kim, ``Empirical measurement of client contribution
  for federated learning with data size diversification,'' \emph{IEEE Access},
  vol.~10, pp. 118\,563--118\,574, 2022, doi:~10.1109/ACCESS.2022.3210950.

\bibitem{balakrishnan2021diverse}
R.~Balakrishnan, T.~Li, T.~Zhou, N.~Himayat, V.~Smith, and J.~Bilmes, ``Diverse
  client selection for federated learning via submodular maximization,'' in
  \emph{International Conference on Learning Representations}, 2021, p.~18,
  url:~\url{https://openreview.net/forum?id=nwKXyFvaUm}.

\bibitem{shen2022fast}
G.~Shen, D.~Gao, D.~Song, X.~Zhou, S.~Pan, W.~Lou, F.~Zhou \emph{et~al.},
  ``Fast heterogeneous federated learning with hybrid client selection,''
  \emph{arXiv preprint arXiv:2208.05135}, 2022.

\bibitem{mnist-2010}
Y.~LeCun, C.~Cortes, and C.~Burges, ``{MNIST} handwritten digit database,'' ATT
  Labs, Tech. Rep., 2010, url:~\url{http://yann.lecun.com/exdb/mnist},.

\bibitem{cifar10}
A.~Krizhevsky, ``Learning multiple layers of features from tiny images,''
  University of Toronto, Tech. Rep., 2009,
  url~\url{https://www.cs.toronto.edu/~kriz/learning-features-2009-TR.pdf},.

\bibitem{netzer2011reading}
Y.~Netzer \emph{et~al.}, ``Reading digits in natural images with unsupervised
  feature learning,'' in \emph{Workshop on Deep Learning and Unsupervised
  Feature Learning (NIPS)}, 2011.

\bibitem{xu2022opv2v}
R.~Xu, H.~Xiang, X.~Xia, X.~Han, J.~Li, and J.~Ma, ``{OPV2V}: An open benchmark
  dataset and fusion pipeline for perception with vehicle-to-vehicle
  communication,'' in \emph{2022 International Conference on Robotics and
  Automation (ICRA)}.\hskip 1em plus 0.5em minus 0.4em\relax IEEE, 2022, pp.
  2583--2589, doi:~10.1109/ICRA46639.2022.9812038.

\bibitem{li2022v2x}
Y.~Li \emph{et~al.}, ``{V2X-Sim}: Multi-agent collaborative perception dataset
  and benchmark for autonomous driving,'' \emph{IEEE Robotics and Automation
  Letters}, vol.~7, no.~4, pp. 10\,914--10\,921, 2022,
  doi:~10.1109/LRA.2022.3192802.

\bibitem{xu2023v2v4real}
R.~Xu \emph{et~al.}, ``V2v4real: A real-world large-scale dataset for
  vehicle-to-vehicle cooperative perception,'' in \emph{The IEEE/CVF Computer
  Vision and Pattern Recognition Conference (CVPR)}, 2023.

\end{thebibliography}
